\UseRawInputEncoding
\documentclass[conference]{IEEEtran}
\usepackage{amsmath,amsfonts}
\usepackage{algorithmic}
\usepackage{algorithm}
\usepackage{array}
\usepackage[caption=false,font=normalsize,labelfont=sf,textfont=sf]{subfig}
\usepackage{textcomp}
\usepackage{stfloats}
\usepackage{url}
\usepackage{verbatim}
\usepackage{graphicx}
\usepackage{cite}

\usepackage{hyperref}
\usepackage{url} 

\usepackage{bm} 
\usepackage{multirow} 
\usepackage{xcolor} 
\usepackage{textcomp} 
\usepackage{diagbox}
\usepackage[table]{xcolor} 
\definecolor{lightgray}{gray}{0.9}
\definecolor{lightblue}{rgb}{0.85,0.92,1.0}
\definecolor{lightred}{rgb}{1.0, 0.85, 0.85}

\begin{document}

\title{Object Identification Under Known Dynamics:\\ A PIRNN Approach for UAV Classification}

\author{\IEEEauthorblockN{1\textsuperscript{st} Nyi Nyi Aung}
\IEEEauthorblockA{\textit{Department of Mechanical and Industrial Engineering} \\
\textit{Louisiana State University}\\
Baton Rouge, USA \\
naung1@lsu.edu}
\and
\IEEEauthorblockN{2\textsuperscript{nd} Neil Muralles}
\IEEEauthorblockA{\textit{Department of Mechanical and Industrial Engineering} \\
\textit{Louisiana State University}\\
Baton Rouge, USA \\
nmural2@lsu.edu}
\and
\IEEEauthorblockN{3\textsuperscript{rd} Adrian Stein}
\IEEEauthorblockA{\textit{Department of Mechanical and Industrial Engineering} \\
\textit{Louisiana State University}\\
Baton Rouge, USA \\
astein@lsu.edu}
}

\maketitle

\begin{abstract}
This work addresses object identification under known dynamics in unmanned aerial vehicle applications, where learning and classification are combined through a physics-informed residual neural network. The proposed framework leverages physics-informed learning for state mapping and state-derivative prediction, while a softmax layer enables multi-class confidence estimation. Quadcopter, fixed-wing, and helicopter aerial vehicles are considered as case studies. The results demonstrate high classification accuracy with reduced training time, offering a promising solution for system identification problems in domains where the underlying dynamics are well understood.
\end{abstract}


\section{INTRODUCTION}
\label{subsec:introduction}
The increasing deployment of unmanned aerial vehicles (UAVs) in civilian and defense sectors has elevated the importance of dynamic modeling and intent inference for tasks such as control, classification, and anomaly detection. Traditional approaches to UAV identification rely primarily on visual, radio-frequency, or pattern-based features, which are vulnerable in contested or adversarial environments \cite{perrusquia_uncovering_2024,perrusquia_uncovering_2025}. To overcome these limitations, there is growing interest in leveraging the known physics of motion as an additional inductive bias for learning and classification.

Physics-Informed Neural Networks (PINNs) have emerged as a promising framework that embeds physical laws, typically in the form of differential equations, directly into the loss function of a neural network \cite{gu_physics-informed_2024,bianchi_physics-informed_2024,jiang_physics-informed_2024}. This approach enables simultaneous learning from data and enforcement of governing equations, making PINNs highly suitable for modeling UAV dynamics where partial physics is known but direct state measurements or labels are limited.

PINNs have been widely applied for UAV state estimation \cite{perrusquia_novel_2024,kamath_physics-informed_2024}, aerodynamic parameter identification \cite{lin_physics-informed_2025}, and robust dynamics modeling in the presence of faults or external disturbances \cite{ma_development_2024,lu_enhanced_2024}. In particular, work \cite{kamath_physics-informed_2024} proposed a visual servoing framework where a PINN learns the relationship between image-plane variations and control inputs without requiring an explicit inverse Jacobian. Other works have demonstrated the effectiveness of PINNs in real-time applications such as adaptive nonlinear MPC \cite{hong_physics-guided_2023,manzoor_model_2023,sanyal_ramp-net_2023}. Other work has been used to estimate the hypersonic vehicle weight using PINN~\cite{chen_estimation_2022}. 
A hybrid PINN-UKF approach has also been proposed to improve state estimation in dynamic systems~\cite{de_curto_hybrid_2024}. These methods are particularly advantageous in systems like UAVs, where capturing full dynamics is difficult due to unmodeled effects such as aerodynamic coupling, actuator delays, and environmental variability \cite{abdulkadirov_physics-aware_2025,yazdannik_nonlinear_2023}.

While most prior work has focused on learning UAV dynamics for control, trajectory tracking, or fault diagnosis, relatively little attention has been given to the inverse problem of identifying the class or type of UAV from its observed motion. In general, learning-based classification algorithms require large datasets to ensure sufficient diversity, yet collecting thousands of trajectories for each UAV type is often impractical. In contrast, PINNs can achieve effective learning with comparatively smaller datasets, as they integrate prior physics knowledge into the training process \cite{raissi_physics-informed_2019}. This property makes them a compelling choice for UAV classification. Motivated by this, we propose a novel ResNet-style PINN framework, termed the Physics-Informed Residual Neural Network (PIRNN), to identify unknown UAVs by leveraging their compliance with known dynamic models.


Our approach assumes a library of known UAV dynamic classes (e.g., quadcopter, fixed wing, and helicopter), on which learning-based classification is performed. Compared to the inherent complexity of UAV dynamics, arising from aerodynamics and flight behavior, both the neural network architecture and dataset used in this work (summarized in Tab.~\ref{tab:pinn_architecture} and Tab.~\ref{tab:hyper}) are relatively small. Nevertheless, by integrating prior physics knowledge within a ResNet-style framework, the proposed method demonstrates that physics-informed learning enables effective multi-class classification with minimal datasets and modest network size, while still achieving high accuracy. This contribution provides strong evidence that physics-informed learning is a viable and efficient approach for UAV classification.
This work builds upon advances in control-physics informed learning \cite{osorio_quero_physics-informed_2024,gu_physics-informed_2024}, estimation \cite{bianchi_physics-informed_2024,na_identification_2024}, and robust adaptive control design \cite{sanyal_ramp-net_2023}, and extends these methods toward a new capability which is object identification under known dynamics. This manuscript is structured as follows: Section~\ref{sec:methodology} provides the methodology, Section~\ref{sec:results} illustrates the results, and Section~\ref{sec:conclusion} concludes the findings. The source code and models used in this study are openly accessible at \href{https://github.com/NyiNyi-14/PINN_based_UAV_Classification.git}{GitHub}. 
%
%
%
\section{METHODOLOGY}
\label{sec:methodology}

Since both learning and classification are integrated for UAV identification, the corresponding workflows are outlined in the following section. The learning process is illustrated in Fig.~\ref{fig:training_flow}, while the classification process is shown in Fig.~\ref{fig:skip_con}~(b). In contrast to conventional data-driven neural networks, PINNs embed known physical principles—commonly expressed as partial differential equations (PDEs) directly into the training objective by incorporating them into the loss function \cite{raissi_physics-informed_2019}. This physics-based integration enables the network to maintain consistency with governing dynamics, thereby enhancing generalization, limiting overfitting, and ensuring closer alignment with real physical behavior \cite{lawal_physics-informed_2022}.

%
%
%
\subsection{Network Architecture}
\begin{figure*}
    \centering
    \includegraphics[clip, trim = 47 570 46 23, width=\linewidth]{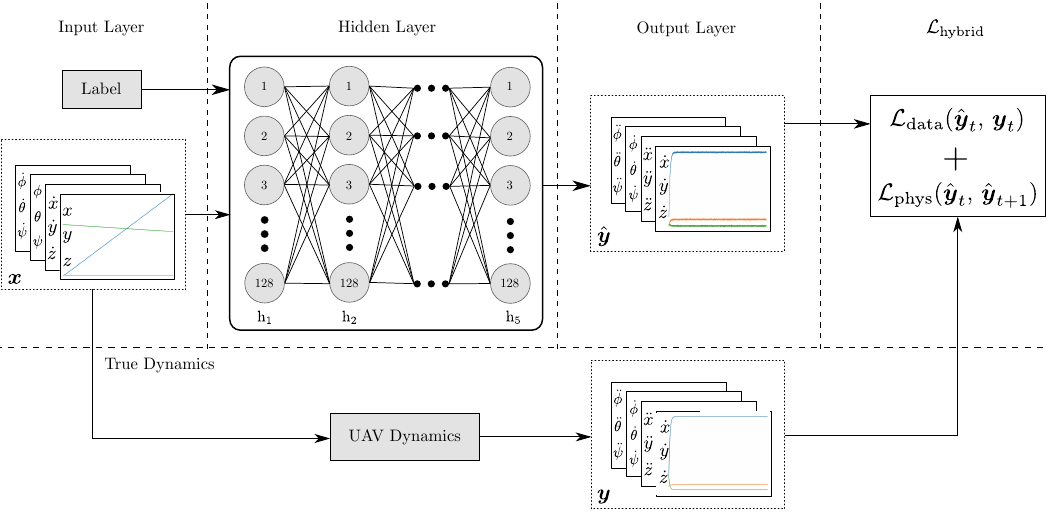}
    \caption{Learning Workflow of the Physics-informed Residual Neural Network (PIRNN).}
    \label{fig:training_flow}
\end{figure*}
\begin{figure*}
    \centering
    \includegraphics[clip, trim = 62 620 75 15, width=\linewidth]{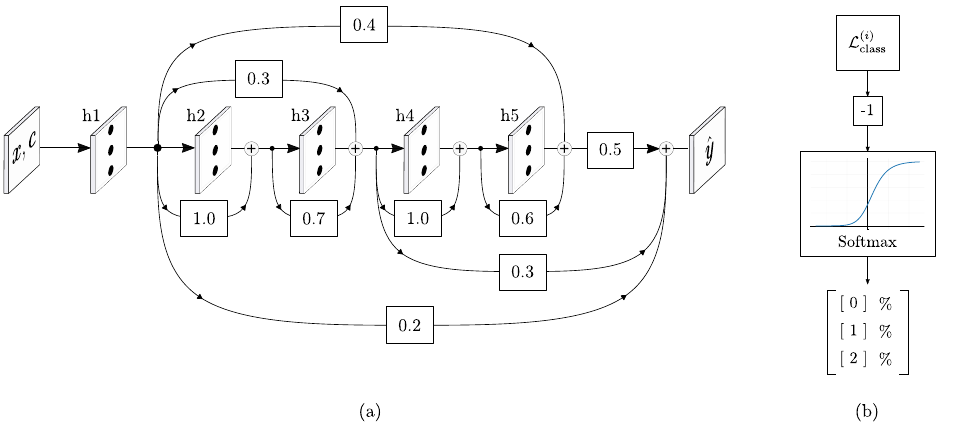}
    \caption{(a) ResNet Style Neural Network Architecture (b) Softmax-weighted Multi-class Confidence Estimation}
    \label{fig:skip_con}
\end{figure*}

As illustrated in Fig.~\ref{fig:training_flow}, the state vector of the UAVs, denoted by $\bm{x}$, along with the class label $\bm{c}$, is provided as input to the neural network. Since this work employs a supervised learning framework, the class label is explicitly included during training. The architecture of the PIRNN is inspired by the ResNet structure, as shown in Fig.~\ref{fig:skip_con} (a), where skip connections are implemented. Although the ResNet structure may appear excessive given the small size of the network used in this work, the inclusion of residual connections with independently learned weights enhances representation learning and mitigates the risk of overfitting. Furthermore, the network is designed to extract diverse features not only from the final layer but also from intermediate layers, enriching the learned representation.

The number of hidden layers, neurons per layer, and activation functions used are summarized in Table~\ref{tab:pinn_architecture}. Although ReLU is commonly adopted in ResNet architectures, this work uses the hyperbolic tangent, $\tanh(\cdot)$ activation function. The choice of $\tanh(\cdot)$ is motivated by its smoothness properties, which are particularly beneficial for predicting derivatives, and its bounded output range [-1, 1], which helps mitigate exploding gradients in physics-based loss terms. Furthermore, $\tanh(\cdot)$ treats positive and negative inputs symmetrically, making it well-suited for capturing the oscillatory behavior of UAVs in roll, pitch, and yaw dynamics.
\begin{table}
\centering
\caption{Selected parameters and functions for PIRNN architecture.}
\renewcommand{\arraystretch}{1.3}
\begin{tabular}{|c|c|l|}
\hline
\textbf{Layer type} & \textbf{Neurons} & \textbf{Activation} \\
\hline
Input  & $\bm{x} + \bm{c}$ & - \\
\hline
Hidden 1 & 128 & $\bm{h}_1 = \tanh(\cdot)$ \\
Hidden 2 & 128 & $\bm{h}_2 = \tanh(\cdot) + \bm{h}_1$ \\
Hidden 3 & 128 & $\bm{h}_3 = \tanh(\cdot) + 0.7 \cdot \bm{h}_2 + 0.3 \cdot \bm{h}_1$ \\
Hidden 4 & 128 & $\bm{h}_4 = \tanh(\cdot) + \bm{h}_3$ \\
Hidden 5 & 128 & $\bm{h}_5 = \tanh(\cdot) + 0.6 \cdot \bm{h}_4 + 0.4 \cdot \bm{h}_1$ \\
\hline
Output & $\bm{\hat{y}}$ & $\mathrm{Linear} + 0.5 \cdot \bm{h}_5 + 0.3 \cdot \bm{h}_3 + 0.2 \cdot \bm{h}_1$ \\
\hline
\end{tabular}
\label{tab:pinn_architecture}
\end{table}
%
%
%
\subsection{Loss Function Design}
Although incorporating PDEs into neural networks is a defining characteristic of physics-informed neural networks, applying this approach to UAV classification presents notable challenges. Different types of UAVs exhibit distinct dynamic behaviors, which would require the formulation and integration of separate physical constraints or PDEs for each class. Embedding these constraints directly into the neural network would significantly increase model complexity, requiring larger architectures and greater computational resources.

To address this, the proposed PIRNN adopts a physics regularization strategy instead of explicitly encoding PDEs. A common physical trait shared across all UAV types is that their state trajectories evolve smoothly over time. Since the classification task in this work is grounded in known system dynamics, this smoothness property is used as a general physical constraint. By penalizing abrupt changes in predicted states across time steps, the model enforces temporal consistency, thereby preserving the physical plausibility of UAV motion across different types. Based on this principle, the hybrid loss function used in this work is defined as:
\begin{align} 
    \label{eq:hybrid_loss}
    \mathcal{L}_{\mathrm{hybrid}} = \lambda_{\mathrm{data}} \ \mathcal{L}_{\mathrm{data}} + \lambda_{\mathrm{phys}} \ \mathcal{L}_{\mathrm{phys}},
\end{align}
where $\lambda_{\mathrm{data}}$ and $\lambda_{\mathrm{phys}}$ represent the weights assigned to the data loss and the physics regularization loss, respectively. In this work, these values are set to $\lambda_{\mathrm{data}} = 1$ and $\lambda_{\mathrm{phys}} = 0.2$. The individual loss components are defined as follows:
\begin{subequations}
    \begin{align}
        {\mathcal{L}}_{\mathrm{data}} = \frac{1}{T} \sum_{t=1}^{T} \left\| \hat{\bm{y}}_t - \bm{y}_t \right\|^2,\\
        \mathcal{L}_{\mathrm{phys}} = \frac{1}{T - 1} \sum_{t=1}^{T-1} \left\| \hat{\bm{y}}_{t+1} - \hat{\bm{y}}_t \right\|^2,
    \end{align}
\end{subequations}
where $\bm{y}_t$ denotes the ground truth derivative of the input state $\bm{x}$, and $\hat{\bm{y}}_t$ represents the predicted derivative produced by the network at time step $t$. The term $T$ denotes the total number of time steps in the trajectory. The working principle of the hybrid loss function is based on training the neural network to approximate the ground truth by minimizing the discrepancy between predictions and actual values, while simultaneously penalizing abrupt changes in state derivatives to enforce physical consistency.
%
%
%
\subsection{Softmax-Weighted Confidence Estimation for Multi-Class Classification}
A physics-based classification approach is employed, and the workflow of the classifier is illustrated in Fig.~\ref{fig:skip_con} (b). First, the trained PIRNN model predicts the derivative states of the given trajectory three times, once for each UAV type, as follows:
\begin{align}
    \hat{\bm{y}}_{t}^{(i)} = f_{\theta}(\bm{x}_t, \bm{c}_i),
\end{align}
where $f_{\theta}$ denotes the trained PIRNN model, $\hat{\bm{y}}_{t}^{(i)}$ is the predicted derivative for class $i$, $\bm{c}_i \in \mathbb{R}^C$ is the one-hot encoded class vector, and ${i} = [0, 1, 2]$ corresponds to quadcopter, fixed-wing, and helicopter, respectively. Each prediction is compared to the ground truth derivative to compute the average loss for class $i$, denoted as $\mathcal{L}_{\mathrm{class}}^{(i)}$:
\begin{align}
    \mathcal{L}_{\mathrm{class}}^{(i)} = \frac{1}{T} \sum_{t=1}^{T} \left\lVert \hat{\bm{y}}_{t}^{(i)} - \bm{y}_t \right\rVert^2.
\end{align}
The resulting loss values are used to compute confidence scores via a softmax function. To ensure that lower loss values correspond to higher confidence scores, the losses are negated before being passed into the softmax operator $\sigma(\cdot)$, yielding a normalized distribution that sums to 1 as follows:
\begin{align}
    p^{(i)} = \sigma(\mathcal{L}^{(i)}_{\mathrm{class}}) = \frac{\exp\left(-\gamma \ \mathcal{L}^{(i)}_{\mathrm{class}} \right)}{\sum_{j=0}^{C} \exp\left(-\gamma \ \mathcal{L}^{(j)}_{\mathrm{class}} \right)},
\end{align}
where $p^{(i)}$ represents the model’s confidence that the input trajectory belongs to class $i$, $C$ is the total number of classes, and $\gamma$ is a temperature-like scaling factor. In this work, $\gamma = 10$ is selected to produce a sharper distribution.
Finally, the UAV class associated with the highest confidence score is selected as the predicted label:
\begin{equation}
    \hat{c} = \min_{i} \mathcal{L}^{(i)},
\end{equation}
where $\hat{c}$ denotes the predicted class.
%
%
%
\subsection{Modeling and Dynamic Characteristics of UAV Types}
\begin{figure*}
    \centering
    \includegraphics[clip, trim = 10 660 10 10, width=\linewidth]{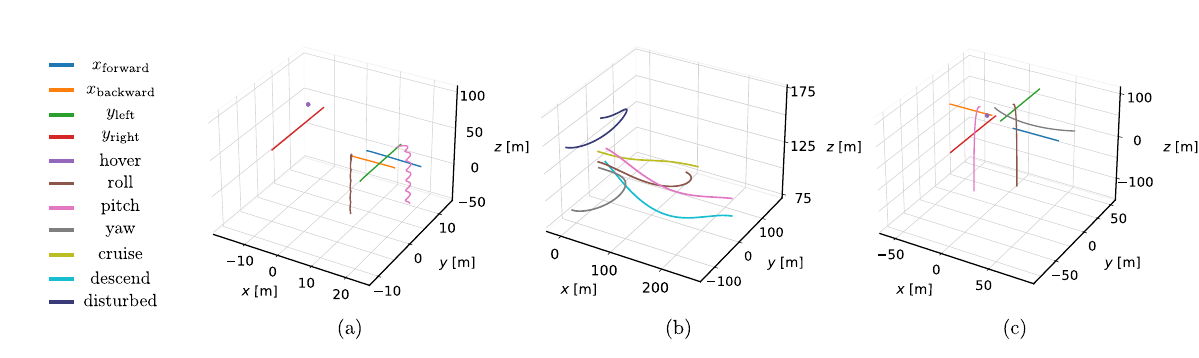}
    \caption{Sample Trajectories of UAVs: (a) Quadcopter, (b) Fixed-wing, (c) Helicopter.}
    \label{fig:UAV_traj}
\end{figure*}

Since the states of UAVs serve as inputs to the PIRNN, it is essential to generate a diverse set of trajectories for each UAV type. This allows the model to learn a wide range of dynamic behaviors and generalize across different flight conditions. To this end, the dynamic characteristics of each UAV are introduced in the following subsections. For each UAV type, 1000 trajectories are simulated over a duration of 10 seconds, capturing a variety of flight maneuvers such as hovering, climbing, rolling, pitching, yawing, as well as disturbed and failure conditions.
Sample trajectories for each UAV type are shown in Fig.~\ref{fig:UAV_traj}. As expected, the fixed-wing UAV exhibits longer trajectories compared to the others, despite having the same simulation time. This is consistent with its inherently higher forward velocity relative to rotary-wing UAVs.
%
%
%
\subsubsection{Quadcopter}
As the name “quadcopter” suggests, it is driven by four electric motors, and a total of 12 states are considered to describe its motion and dynamics as follows \cite{abdulkareem_modeling_2022}:
\begin{align} \label{eq:quad_state}
    \bm{x}_{\mathrm{quad}} =
    \left[
    x,\, y,\, z,\, \dot{x},\, \dot{y},\, \dot{z},\, \phi,\, \theta,\, \psi,\, p,\, q,\, r
    \right] \in \mathbb{R}^{12}.
\end{align}
Here, $\bm{\xi} = (x, y, z) \in \mathbb{R}^3$ represents the position of the quadcopter’s center of mass in an inertial reference frame, and the corresponding linear velocity is defined as \mbox{$\bm{v} = (\dot{x}, \dot{y}, \dot{z}) \in \mathbb{R}^3$}. The orientation is described by the Euler angles \mbox{$\bm{\eta} = (\phi, \theta, \psi) \in \mathbb{R}^3$}, where $\phi$, $\theta$, and $\psi$ denote the roll, pitch, and yaw angles, respectively, with respect to the body frame. The body angular velocity is denoted by $\bm{\omega} = (p, q, r) \in \mathbb{R}^3$. Since Euler angles are defined in the body frame, while the position is defined in the inertial frame, a rotation matrix $\bm{R}(\phi, \theta, \psi)$ is introduced to relate the two frames as:
\begin{equation}
\bm{R}(\phi, \theta, \psi) = \bm{R}_x(\phi) \ \bm{R}_y(\theta) \ \bm{R}_z(\psi),
\end{equation}
\begin{equation}
\begin{aligned}
\bm{R}_x(\phi) &=
\begin{bmatrix}
1 & 0 & 0 \\
0 & c_\phi & -s_\phi \\
0 & s_\phi & c_\phi
\end{bmatrix};\quad
\bm{R}_y(\theta) =
\begin{bmatrix}
c_\theta & 0 & s_\theta \\
0 & 1 & 0 \\
-s_\theta & 0 & c_\theta
\end{bmatrix};\\
\bm{R}_z(\psi) &=
\begin{bmatrix}
c_\psi & -s_\psi & 0 \\
s_\psi & c_\psi & 0 \\
0 & 0 & 1
\end{bmatrix}.
\end{aligned}
\end{equation}
Additionally, the transformation matrix $\bm{W}(\phi, \theta)$ is used to convert the angular velocity in the body frame $\bm{\omega} = [p, q, r]^\top$ to the Euler angle rates $\dot{\bm{\eta}} = [\dot{\phi}, \dot{\theta}, \dot{\psi}]^\top$, accounting for their nonlinear relationship as:
\begin{align} 
    \label{eq:E_rate}
    \dot{\bm{\eta}} =
    \mathbf{W}(\phi, \theta) \cdot \bm{\omega}; \quad
    \mathbf{W}(\phi, \theta) =
    \begin{bmatrix}
        1 & s_\phi \, t_\theta & c_\phi \, t_\theta \\
        0 & c_\phi & -s_\phi \\
        0 & \frac{s_\phi}{c_\theta} & \frac{c_\phi}{c_\theta}
    \end{bmatrix},
\end{align}
where $c_{(\cdot)} = \cos(\cdot)$, $s_{(\cdot)} = \sin(\cdot)$ and $t_{(\cdot)} = \tan(\cdot)$ are used for notational brevity. The quadcopter is actuated through the angular velocities of its four motors, each producing both thrust and drag torque. By varying the motor speeds, the vehicle can generate net thrust, roll, pitch, and yaw torques as \cite{abdulkareem_modeling_2022}:
\begin{align}
    T_i = k_\mathrm{T} \omega_i^2; \quad \tau_i = k_\mathrm{D} \omega_i^2; \quad i = [1,2,3,4],
\end{align}
where $k_\mathrm{T}$ and $k_\mathrm{D}$ denote the thrust and drag torque coefficients, respectively. $T_i$ and $\tau_i$ represent the thrust and drag torque generated by the $i$-th rotor’s angular velocity $\omega_i$. The overall control input vector $\bm{u}$ is then defined as:
\begin{align}
    \begin{bmatrix}
    u_1 \\
    u_2 \\
    u_3 \\
    u_4
    \end{bmatrix}
    =
    \begin{bmatrix}
    1 & 1 & 1 & 1 \\
    0 & -l & 0 & l \\
    l & 0 & -l & 0 \\
    0 & 0 & 0 & 0
    \end{bmatrix}
    \begin{bmatrix}
    T_1 \\
    T_2 \\
    T_3 \\
    T_4
    \end{bmatrix}
    +
    \begin{bmatrix}
    0 \\
    0 \\
    0 \\
    \tau_1 - \tau_2 + \tau_3 - \tau_4
    \end{bmatrix}.
\end{align}
From this, the net thrust $\bm{f}_{\mathrm{thrust}}$ and body torque $\bm{\tau}_{\mathrm{body}}$ are given by:
\begin{align}
    \bm{f}_{\mathrm{thrust}} = \begin{bmatrix} 0 \\ 0 \\ u_1 \end{bmatrix}; \quad 
\bm{\tau}_{\mathrm{body}} = \begin{bmatrix} u_2 \\ u_3 \\ u_4 \end{bmatrix}.
\end{align}
Then, the translational and rotational accelerations are computed as:
\begin{align} 
    \dot{\bm{v}} = \frac{1}{m} \left( \bm{R}(\phi, \theta, \psi) \cdot \bm{f}_{\mathrm{thrust}} - k_\mathrm{drag, v} \cdot \bm{v} \right)
    + \begin{bmatrix} 0 \\ 0 \\ -g \end{bmatrix}, \label{eq:T_acc}\\ 
        \bm{\dot{\omega}} = \bm{I}^{-1} \left( \bm{\tau}_{\mathrm{body}} - k_\mathrm{drag, \omega} \cdot  \bm{\omega} - \bm{\omega} \times (\bm{I} \cdot \bm{\omega}) \right) \label{eq:R_acc}, 
\end{align}
where, $k_\mathrm{drag, v}$ and $k_\mathrm{drag, \omega}$ represent the linear and angular drag coefficient. The complete quadcopter dynamics employed in this work are modeled using the Eqs.~\eqref{eq:quad_state}-\eqref{eq:R_acc}. Sample trajectories generated from these dynamics are illustrated in Fig.~\ref{fig:UAV_traj} (a).
%
%
%
\subsubsection{Fixed-Wing UAV}
Similar to the quadcopter, the fixed-wing UAV also utilizes a 12-dimensional state vector, but the variables are arranged in a different order as defined in \cite{beard_small_2012}:
\begin{align} 
\label{eq:fw_state}
    \bm{x}_{\mathrm{fw}} =
    \left[
    \dot{x},\, \dot{y},\, \dot{z},\,
    p,\, q,\, r, \,
    \phi,\, \theta,\, \psi,\, 
    x,\, y,\, z
    \right] \in \mathbb{R}^{12}.
\end{align}
The control input vector is defined as $\bm{u} = [\delta_\mathrm{t}, \delta_\mathrm{a}, \delta_\mathrm{e}, \delta_\mathrm{r}]$, representing the throttle and the three primary control surfaces: aileron, elevator, and rudder, which are responsible for roll, pitch, and yaw control, respectively. In this work, the fixed-wing UAV is modeled under the assumption that the dominant forces acting on it include gravitational force $\bm{f}_\mathrm{g}$, aerodynamic force $\bm{f}_\mathrm{a}$, and propulsion force $\bm{f}_\mathrm{p}$. The gravitational force acts on the center of mass in the inertial frame and is transformed into the body frame using the rotation matrix, resulting in:
\begin{align}
    \bm{f}_\mathrm{g}^{\mathbb{B}} = 
    \begin{bmatrix}
        f_\mathrm{x, g} \\
    f_\mathrm{y, g} \\
    f_\mathrm{z, g} 
    \end{bmatrix}
    = \bm{R}^{\mathbb{B}}(\phi, \theta) \cdot
    \begin{bmatrix}
    0 \\
    0 \\
    mg
    \end{bmatrix}
    =
    \begin{bmatrix}
    - s_\theta \\
    c_\theta s_\phi \\
    c_\theta c_\phi
    \end{bmatrix} \cdot
    mg,
\end{align}
where $\bm{R}^{\mathbb{B}}(\phi, \theta)$ is the rotation matrix that transforms vectors from the inertial frame to the body frame. The aerodynamic force can be decomposed into longitudinal and lateral components. The longitudinal aerodynamic force is defined as:
\begin{align}
    \begin{bmatrix}
    f_\mathrm{x, aero} \\
    f_\mathrm{z, aero}
    \end{bmatrix}
    =
    \begin{bmatrix}
    c_\alpha & -s_\alpha \\
    s_\alpha & c_\alpha
    \end{bmatrix}
    \begin{bmatrix}
    - F_{\mathrm{drag}} \\
    - F_{\mathrm{lift}}
    \end{bmatrix},
\end{align}
where $\alpha = \tan^{-1}\left(\frac{\dot{z}}{\dot{x}}\right)$ represent the angle of attack. The drag and lift forces can then be expressed as \cite{beard_small_2012}:
\begin{align}
    F_{\text{lift}} = \frac{1}{2} \rho V_\mathrm{a}^2 S C_\mathrm{L}(\alpha, \delta_\mathrm{e}); \:
    F_{\mathrm{drag}} = \frac{1}{2} \rho V_\mathrm{a}^2 S  C_\mathrm{D}(\alpha, \delta_\mathrm{e}).
\end{align}
Here, $\rho$ denotes the air density, $V_\mathrm{a} = \sqrt{\dot{x}^2 + \dot{y}^2 + \dot{z}^2}$ represents the airspeed, and $S$ is the wing area of the fixed-wing UAV. The coefficients $C_\mathrm{L}(\alpha, \delta_\mathrm{e})$ and $C_\mathrm{D}(\alpha, \delta_\mathrm{e})$ are the lift and drag coefficients, respectively, expressed as functions of the angle of attack $\alpha$ and elevator deflection $\delta_\mathrm{e}$. 
The lateral aerodynamic force is given by:
\begin{align}
    f_{\mathrm{y,aero}} = \frac{1}{2} \rho V_\mathrm{a}^2 S C_\mathrm{y}(\beta, \delta_\mathrm{r}),
\end{align}
where $C_\mathrm{y}(\beta, \delta_\mathrm{r})$ is the non-dimensional aerodynamic coefficient as functions of the sideslip angle $\beta$, and the rudder deflection $\delta_\mathrm{e}$. The sideslip angle is computed as $\beta = \sin^{-1}\left( \frac{\dot{y}}{V_\mathrm{a}} \right)$.
As for the propulsion force, it originates from the thrust and torque generated by the motor–propeller pair. In this work, the propulsive force is modeled as:
\begin{align}
    \omega_\mathrm{prop} = \delta_t \cdot \omega_\mathrm{in}; \quad
    f_{x,\mathrm{prop}} = \frac{\rho D^4}{4\pi^2} \omega_\mathrm{prop}^2 C_\mathrm{t},
\end{align}
where $\delta_t \in [0, 1]$ is the throttle input, $D$ is the propeller diameter, $C_\mathrm{t}$ is the thrust coefficient, and $\omega_\mathrm{prop}$ is the actual angular velocity of the propeller, determined by the input speed $\omega_\mathrm{in}$ corresponding to the applied throttle setting.

The total force acting on the fixed-wing UAV is obtained by combining the forces from all three sources, and is given by:
\begin{align}
    \begin{bmatrix}
    F_\mathrm{x} \\
    F_\mathrm{y} \\
    F_\mathrm{z}
    \end{bmatrix}
    =
    \begin{bmatrix}
    f_{\mathrm{x,aero}} \\
    f_{\mathrm{y,aero}} \\
    f_{\mathrm{z,aero}}
    \end{bmatrix}
    +
    \begin{bmatrix}
    f_{\mathrm{x,prop}} \\
    0 \\
    0
    \end{bmatrix}
    +
    \begin{bmatrix}
    f_{\mathrm{x,g}} \\
    f_{\mathrm{y,g}} \\
    f_{\mathrm{z,g}}
    \end{bmatrix}.
\end{align}
Then, the moments for roll, pitch, and yaw are defined as \cite{beard_small_2012}:
\begin{align}
    \bm{\tau} = 
    \begin{bmatrix}
    \tau_\mathrm{x} \\[0.3em]
    \tau_\mathrm{y} \\[0.3em]
    \tau_\mathrm{z}
    \end{bmatrix} = 
    \begin{bmatrix}
        \frac{1}{2} \rho V_\mathrm{a}^2 \, S \, b \, C_\mathrm{l}(\beta, \delta_\mathrm{a}) \\[0.3em]
        \frac{1}{2} \rho V_\mathrm{a}^2 \, S \, c \, C_\mathrm{m}(\alpha, \delta_\mathrm{e}) \\[0.3em]
        \frac{1}{2} \rho V_\mathrm{a}^2 \, S \, b \, C_\mathrm{n}(\beta, \delta_\mathrm{r})
    \end{bmatrix},
\end{align}
where $b$ is the wingspan, $c$ is the mean aerodynamic chord, and $C_\mathrm{l}(\beta, \delta_\mathrm{a})$, $C_\mathrm{m}(\alpha, \delta_\mathrm{e})$, and $C_\mathrm{n}(\beta, \delta_\mathrm{r})$ are the aerodynamic moment coefficients about roll, pitch, and yaw, respectively, expressed as functions of the angle of attack $\alpha$, sideslip angle $\beta$, and the corresponding control surface deflections.
After the necessary forces and moments are defined, the translational and rotational accelerations are given by:
\begin{align} 
    \bm{\dot{v}} = 
    \begin{bmatrix}
    \ddot{x} \\[0.3em]
    \ddot{y} \\[0.3em]
    \ddot{z}
    \end{bmatrix}
    =
    \frac{1}{m}
    \begin{bmatrix}
    F_\mathrm{x} \\[0.3em]
    F_\mathrm{y} \\[0.3em]
    F_\mathrm{z}
    \end{bmatrix}
    +
    \begin{bmatrix}
    - q \dot{z} + r \dot{y} \\[0.3em]
    - r \dot{x} + p \dot{z} \\[0.3em]
    - p \dot{y} + q \dot{x}
    \end{bmatrix}, \label{eq:fw_T_acc}\\
        \bm{\dot{\omega}} =
    \begin{bmatrix}
    \dot{p} \\ \dot{q} \\ \dot{r}
    \end{bmatrix} =
     \bm{I}^{-1} \left( \bm{\tau} - \bm{\omega} \times (\bm{I} \cdot \bm{\omega}) \right) \label{eq:fw_R_acc}.
\end{align} 
Then, the Euler angle rate is written as:
\begin{align} 
\label{eq:fw_E_rate}
    \begin{bmatrix}
    \dot{\phi} \\[0.3em]
    \dot{\theta} \\[0.3em]
    \dot{\psi}
    \end{bmatrix}
    =
    \begin{bmatrix}
    1 & t_\theta\, s_\phi & t_\theta\, c_\phi \\[0.3em]
    0 & c_\phi & -s_\phi \\[0.3em]
    0 & \frac{s_\phi}{c_\theta} & \frac{c_\phi}{c_\theta}
    \end{bmatrix}
    \begin{bmatrix}
    p \\[0.3em]
    q \\[0.3em]
    r
    \end{bmatrix},
\end{align}
while the position rate in the inertial frame is defined as:
\begin{equation} \label{eq:fw_P_rate}
    \bm{v}^\mathbb{I} = \bm{R}(\phi, \theta, \psi) \cdot \bm{v}.
\end{equation}
Finally, the fixed-wing UAV is modeled using the Eqs.~\eqref{eq:fw_state}-\eqref{eq:fw_P_rate} and the trajectories generated from this dynamics are shown in Fig.~\ref{fig:UAV_traj} (b).
%
%
%
\subsubsection{Helicopter}
Like the quadcopter model, helicopter uses the same 12 states, given as \cite{budiyono_first_2008}:
\begin{align} 
    \label{eq:heli_state}
    \bm{x}_{\mathrm{heli}} =
    \left[
    x,\, y,\, z,\, \dot{x},\, \dot{y},\, \dot{z},\, \phi,\, \theta,\, \psi,\, p,\, q,\, r
    \right] \in \mathbb{R}^{12}.
\end{align}
The helicopter is controlled by four primary inputs: main rotor thrust $T_\mathrm{m}$, rear rotor thrust $T_\mathrm{r}$, cyclic roll angle $\delta_{\phi}$, and cyclic pitch angle $\delta_{\theta}$. These control inputs can be written as:
\begin{align}
    \begin{bmatrix}
    u_1 \\
    u_2 \\
    u_3 \\
    u_4
    \end{bmatrix}
    =
    \begin{bmatrix}
    T_m \\
    T_m \cdot l \cdot \sin(\delta_\phi) \\
    T_m \cdot l \cdot \sin(\delta_\theta) \\
    T_r \cdot d
    \end{bmatrix}.
\end{align}
Here, $u_1$ represents the main thrust, while $u_2$, $u_3$, and $u_4$ correspond to the roll, pitch, and yaw torques, respectively. Moreover, the helicopter's unique configuration includes a moment arm length $l$, which denotes the horizontal distance from the main rotor thrust to the helicopter’s center of mass, and a main-to-rear rotor distance $d$, representing the horizontal distance between the main and rear rotors. Then, the force and torque vectors are defined as:
\begin{align}
    \bm{f}_\mathrm{thrust} = 
    \begin{bmatrix}
    0 \\
    0 \\
    u_1
    \end{bmatrix}; \quad
    \bm{\tau}_\mathrm{body} = 
    \begin{bmatrix}
    u_2 \\
    u_3 \\
    u_4
    \end{bmatrix}.
\end{align}
For the purposes of this simulation, the rear rotor thrust and the horizontal components of the main rotor thrust in the body frame are considered negligible with respect to direct translational motion. Therefore, the equations for translational and rotational accelerations are assumed to follow the same form as those defined in the quadcopter dynamics, as given in \eqref{eq:T_acc} and \eqref{eq:R_acc}. Based on this assumption, the helicopter UAV is modeled accordingly in this work, and the trajectories generated for the helicopter can be seen in Fig.~\ref{fig:UAV_traj}~(c).
%
%
\subsection{Training the PIRNN}
To accelerate training, a skip connection is integrated into the network architecture, as illustrated in Fig.~\ref{fig:skip_con} (a). Additionally, the training process is conducted on a GPU rather than a CPU to further improve computational efficiency. The specifications of the workstation and the selected hyper-parameters are listed in Table~\ref{tab:work_station} and Table~\ref{tab:hyper}, respectively. It is important to note that the hyper-parameters used in this work were chosen through trial and error; no formal hyper-parameter optimization was performed.

\begin{table}
\centering
\caption{Workstation specifications} \label{tab:work_station}
\renewcommand{\arraystretch}{1.3}
\begin{tabular}{|l|l|}
\hline
\textbf{Specification} & \textbf{Details} \\
\hline
OS & Windows 11 Home \\
GPU & NVIDIA GeForce RTX 4060 Ti, 16.0 GB \\
Processor & AMD Ryzen 9 7950X 16-Core Processor @ 4.50 GHz \\
 RAM & 64.0 GB \\
\hline
\end{tabular}
\end{table}

\begin{figure}
    \centering
    \includegraphics[clip, trim = 4 685 222 10, width=\linewidth]{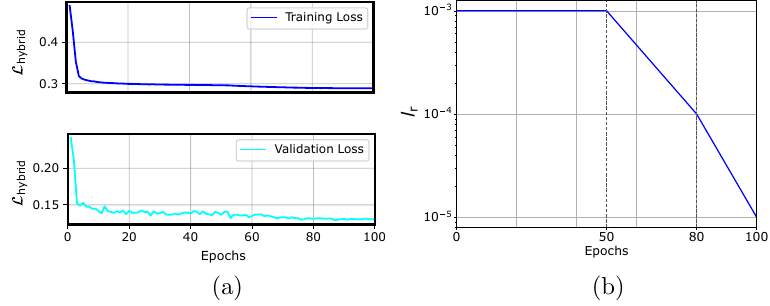}
    \caption{(a) Training Loss vs. Validation Loss of the PIRNN, (b) Learning Rate Scheduling}
    \label{fig:loss_lr}
\end{figure}

\begin{table}
\centering
\caption{Selected hyper-parameters set for PIRNN} \label{tab:hyper}
\renewcommand{\arraystretch}{1.3}
\begin{tabular}{|l|l|l|}
\hline
\textbf{Symbol} & \textbf{Description} & \textbf{Selected set} \\
\hline
$\bm{c}$ & class/ label & [0, 1, 2] \\
$\bm{x}$ & UAV states & [3000 $\cdot$ 1000, 12] \\
$\bm{x}_\mathrm{train}$ & training set & 80\% $\cdot$ $\bm{x}$ \\
$\bm{x}_\mathrm{val}$ & validating set & 10\% $\cdot$ $\bm{x}$ \\
$\bm{x}_\mathrm{test}$ & testing set & 10\% $\cdot$ $\bm{x}$ \\
$\mathcal{D}_\mathrm{train}$ & training batch size & 256 \\
$\mathcal{D}_\mathrm{val,test}$ & validating/ testing batch size & $1 \cdot 10^{3}$ \\
$N_\mathrm{e}$ & number of epochs & 100 \\
$N_\mathrm{patience}$ & early stopping patience & 30 \\
$\delta_\mathrm{loss}$ & early stopping threshold & $1 \cdot 10^{-5}$ \\
$l_\mathrm{r}$ & learning rate & see Fig.~\ref{fig:loss_lr} (b) \\
$t_\mathrm{duration}$ & training duration & 55 mins 37 secs \\
\hline
\end{tabular}
\end{table}
As shown in Fig.~\ref{fig:loss_lr} (b), a scheduled learning rate $l_\mathrm{r}$, is used in this work instead of a constant learning rate. In the initial learning phase (Phase~1), a relatively high learning rate is employed to allow the loss function to explore the solution space and converge rapidly. After half of the total training epochs, the model enters an intermediate learning phase (Phase~2), during which $l_\mathrm{r}$ is reduced by an order of magnitude to strike a balance between convergence speed and stability. Finally, in the late training phase (Phase~3), the learning rate is decreased by another order of magnitude to ensure that the model converges smoothly without overshooting the minima.
%
%
%
\section{RESULTS}
\label{sec:results}
After training the network for 100 epochs, the training and validation losses are plotted in Fig.~\ref{fig:loss_lr} (a) to evaluate model performance and computational efficiency. As shown, the training loss begins to plateau after approximately 40 epochs, while the validation loss stabilizes even earlier, around epoch 20. Although both losses continue to decrease, the rate of improvement becomes marginal beyond these points. While the performance could likely be improved with a more rigorously optimized hyperparameter set and extended training time, the current configuration is considered sufficient for the purposes of this study, as reflected in the loss curves in Fig.~\ref{fig:loss_lr}~(a).
%
%
%
\subsection{Performance Evaluation of the Proposed Algorithm}
\begin{figure}
    \centering
    \includegraphics[clip, trim = 5 545 20 10, width=\linewidth]{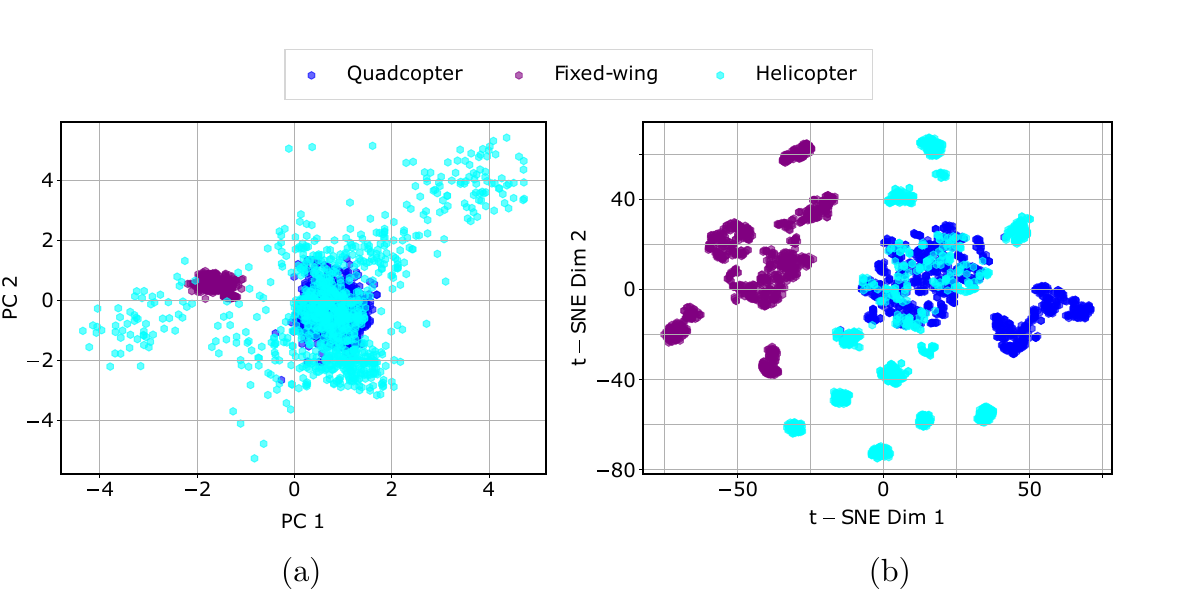}
    \caption{Data distribution of UAVs: (a) Principal Component Analysis (PCA), (b) t-distributed Stochastic Neighbor Embedding (t-SNE).}
    \label{fig:PCA_tSNE}
\end{figure}

The algorithm is evaluated using 300 unseen trajectories for classification, and a subset of the results is presented in Table~\ref{tab:sample_results}. As shown, the algorithm achieves correct predictions for 11 out of the 12 test samples, with only one misclassification, highlighted in light red. While the overall performance is promising, the classifier demonstrates some difficulty in distinguishing between the quadcopter and helicopter, as indicated by the light blue highlights. Although the predictions in these cases are correct, the confidence scores between the two classes are often very close, sometimes around 51\% and 49\%, indicating a degree of uncertainty. This behavior is expected, as quadcopters and helicopters share similar flight characteristics in hovering, climbing, and descending scenarios. To further investigate the overlap in feature space, principal component analysis (PCA) and t-distributed stochastic neighbor embedding (t-SNE) are applied to visualize the data distribution among the UAV types considered in this study. The resulting visualizations are shown in Fig.~\ref{fig:PCA_tSNE}.
\begin{table}
\centering
\renewcommand{\arraystretch}{1.3}
\caption{Classification results showing confidence percentages, prediction, and correctness} \label{tab:sample_results}
\begin{tabular}{|c|c|c|c|c|c|}
\hline
\multirow{2}{*}{\textbf{True Label}} & \multicolumn{3}{c|}{\textbf{Confidence (\%)}} & \multirow{2}{*}{\textbf{Prediction}} & \multirow{2}{*}{\textbf{Correctness}} \\ \cline{2-4}
 & \textbf{Quad} & \textbf{FW} & \textbf{Heli} &  & \\
\hline
\rowcolor{lightblue}
Helicopter   & 49.5  & 0.0   & 50.5  & Helicopter   & True  \\
Helicopter   & 0.0   & 0.0   & 100.0 & Helicopter   & True  \\
\rowcolor{lightblue}
Quadcopter   & 51.37 & 0.0   & 48.63 & Quadcopter   & True  \\
\rowcolor{lightblue}
Helicopter   & 39.26 & 0.01  & 60.73 & Helicopter   & True  \\
\rowcolor{lightblue}
Quadcopter   & 52.43 & 0.0   & 47.57 & Quadcopter   & True  \\
\rowcolor{lightred}
Helicopter   & 50.11 & 0.0   & 49.89 & Quadcopter   & False \\
Fixed-wing   & 0.06  & 99.94 & 0.0   & Fixed-wing   & True  \\
Fixed-wing   & 0.0   & 100.0 & 0.0   & Fixed-wing   & True  \\
Helicopter   & 0.0   & 0.0   & 100.0 & Helicopter   & True  \\
Helicopter   & 0.0   & 0.0   & 100.0 & Helicopter   & True  \\
\rowcolor{lightblue}
Quadcopter   & 52.88 & 0.0   & 47.12 & Quadcopter   & True  \\
Fixed-wing   & 0.0   & 100.0 & 0.0   & Fixed-wing   & True  \\
\hline
\end{tabular}
\end{table}

Generally, PCA maximizes variance by finding directions where the data spreads the most through linear transformations, while t-SNE preserves local similarities using a nonlinear probabilistic approach. As shown in Fig.~\ref{fig:PCA_tSNE}, the fixed-wing UAV exhibits distinct features and spreads clearly in both PCA and t-SNE plots. In contrast, some quadcopter and helicopter samples form overlapping clusters, making it more challenging for the classifier to differentiate between them. Despite this overlap in data distribution, the proposed PIRNN classifier demonstrates excellent performance in UAV classification. The classification report in Table~\ref{tab:classification_report} highlights the model's strong prediction capability, with an overall classification accuracy of 96\%.
\begin{table}
\centering
\renewcommand{\arraystretch}{1.3}
\caption{Classification Report: Precision, Recall, F1-Score, and Support per Class}
\begin{tabular}{|l|c|c|c|c|}
\hline
\textbf{Class} & \textbf{Precision} & \textbf{Recall} & \textbf{F1-score} & \textbf{Support} \\
\hline
Fixed-wing   & 1.00 & 1.00 & 1.00 & 99  \\
Helicopter   & 1.00 & 0.89 & 0.94 & 101 \\
Quadcopter   & 0.90 & 1.00 & 0.95 & 100 \\
\hline
Accuracy & - & - & 0.96 & 300 \\
\hline
\end{tabular}
\label{tab:classification_report}
\end{table}
%
%
%
\subsection{Robustness Analysis of the Proposed Algorithm}
As the algorithm demonstrates high accuracy under ideal conditions, it is of interest to evaluate its robustness. The neural network is trained on clean data and initially tested on simulated trajectories, which represent an idealized scenario. To better reflect real-world conditions and assess robustness, varying levels of Gaussian noise are added to the trajectories. Specifically, zero-mean Gaussian noise ($\mu = 0$) with different standard deviations ($\sigma_{(\cdot)}$) is applied to both the trajectory data and its derivative states, as follows:
\begin{subequations}
    \begin{align}
        \bm{x}_\mathrm{noisy} = \bm{x} + \mathcal{N}(\mu, \sigma_{\mathrm{x}});        \:\sigma_{\mathrm{x}} \in \{3\%, 5\%, 10\%, 15\%\}\\
        \bm{\dot{x}}_\mathrm{noisy} = \bm{\dot{x}} + \mathcal{N}(\mu, \sigma_{\dot{\mathrm{x}}}); \:\sigma_{\dot{\mathrm{x}}} \in \{5\%, 10\%, 15\%, 20\%\}.
    \end{align}
\end{subequations}
The classifier is tested using the aforementioned noisy data, and the corresponding classification report is presented in Table~\ref{tab:noise_classify}, where the robustness of the algorithm is evaluated. An accuracy threshold of 75\% is considered to determine whether the classifier performs reliably under noise or fails the task. As shown in Table~\ref{tab:noise_classify}, the classifier maintains an accuracy above 75\%, as highlighted in light blue, even when both the trajectories and their derivative states are corrupted with Gaussian noise of up to 10\% and 15\% standard deviation.

However, when the noise level exceeds this range, the classification accuracy drops to approximately 60\%, as highlighted in light red, leading to the misclassification of nearly half of the test trajectories. Despite this decline at higher noise levels, the classifier can still be considered robust, as it performs reliably under moderate noise conditions and remains effective even in the presence of overlapping behaviors between quadcopters and helicopters.
\begin{table}
\centering
\renewcommand{\arraystretch}{1.3}
\caption{Classification performance in the presence of trajectory noise}
\begin{tabular}{|c|l|c|c|c|c|}
\hline
$\sigma_{\mathrm{x}}, \ \sigma_{\dot{\mathrm{x}}}$ & \textbf{Class} & \textbf{Precision} & \textbf{Recall} & \textbf{F1-score} & \textbf{Support} \\
\hline
\multirow{4}{*}{\shortstack{3\%, \\5\%}}
  & Quadcopter   & 0.70 & 1.00 & 0.83 & 100.00 \\
  & Fixed-wing   & 1.00 & 1.00 & 1.00 & 99.00 \\
  & Helicopter   & 1.00 & 0.58 & 0.74 & 101.00 \\  \cline{2-6}
    \multicolumn{1}{|c|}{} 
  & \cellcolor{lightblue} Accuracy 
  & \cellcolor{lightblue}       & \cellcolor{lightblue}       & \cellcolor{lightblue} 0.86 
  & \cellcolor{lightblue} 300.00 \\
\hline
\multirow{4}{*}{\shortstack{5\%, \\10\%}}
  & Quadcopter   & 0.63 & 1.00 & 0.77 & 100.00 \\
  & Fixed-wing   & 0.99 & 1.00 & 0.99 & 99.00 \\
  & Helicopter   & 1.00 & 0.41 & 0.58 & 101.00 \\ \cline{2-6}
  \multicolumn{1}{|c|}{} 
  &  \cellcolor{lightblue} Accuracy     &  \cellcolor{lightblue}
  &  \cellcolor{lightblue}     &  \cellcolor{lightblue} 0.80 &  \cellcolor{lightblue} 300.00 \\ 
\hline
\multirow{4}{*}{\shortstack{10\%, \\15\%}}
  & Quadcopter   & 0.60 & 1.00 & 0.75 & 100.00 \\
  & Fixed-wing   & 0.96 & 1.00 & 0.98 & 99.00 \\
  & Helicopter   & 1.00 & 0.31 & 0.47 & 101.00 \\ \cline{2-6}
  &  \cellcolor{lightblue} Accuracy     &  \cellcolor{lightblue}     & 
   \cellcolor{lightblue} &  \cellcolor{lightblue} 0.77 &  \cellcolor{lightblue} 300.00 \\ 
\hline
\multirow{4}{*}{\shortstack{15\%, \\20\%}}
  & Quadcopter   & 0.46 & 1.00 & 0.63 & 100.00 \\
  & Fixed-wing   & 0.91 & 0.53 & 0.67 & 99.00 \\
  & Helicopter   & 1.00 & 0.27 & 0.42 & 101.00 \\ \cline{2-6}
  &  \cellcolor{lightred} Accuracy     &  \cellcolor{lightred}      &  \cellcolor{lightred}
  &  \cellcolor{lightred} 0.60 &  \cellcolor{lightred} 300.00 \\ 
\hline
\end{tabular}
\label{tab:noise_classify}
\end{table}
%
%
%
\section{CONCLUSION}
\label{sec:conclusion}
Supervised learning for object classification is integrated in this work through a ResNet-based PINN, where a hybrid loss function is employed. As discussed, the ResNet-style architecture accelerates training, while the integrated physics-based loss enhances both performance and robustness. It could be shown that the algorithm performs well with just 100 training epochs and a training time of 55 minutes. Since hyper-parameters in this study were selected through trial and error, there is a significant potential for further improvement through systematic hyperparameter optimization. This work serves as a preliminary step toward future research, with plans to adapt and deploy the algorithm in real-world UAV classification scenarios using high-resolution and high-speed camera data.
\bibliographystyle{IEEEtran}
\bibliography{references}  

\end{document}